\pdfoutput=1
\documentclass[11pt]{article}

\usepackage{tikz}
\usetikzlibrary{trees}
\usepackage[preprint]{acl}

\usepackage{times}
\usepackage{latexsym}
\usetikzlibrary{trees}
\usepackage[edges]{forest}
\usepackage{amsmath}
\usepackage{graphicx}
\usepackage{hyperref}
\definecolor{grey}{rgb}{0.5, 0.5, 0.5}
\definecolor{hidden-pink}{rgb}{0,0,0}
\definecolor{hidden-draw}{rgb}{0,0,0}
\tikzstyle{my-box}=[
    rectangle,
    draw=hidden-draw,
    rounded corners,
    text opacity=1,
    minimum height=1.5em,
    minimum width=5em,
    inner sep=2pt,
    align=center,
    fill opacity=.5,
    line width=0.8pt,
]
\tikzstyle{leaf}=[my-box, minimum height=1.5em,
    fill=hidden-pink!80, text=black, align=left, font=\normalsize,
    inner xsep=2pt,
    inner ysep=4pt,
    line width=0.8pt,
]

\usepackage[T1]{fontenc}
\usepackage[utf8]{inputenc}

\usepackage{microtype}

\usepackage{inconsolata}

\usepackage{graphicx}

\title{A Survey of Uncertainty Estimation in LLMs: Theory Meets Practice}
\author{Hsiu-Yuan Huang$^{1,2}$,
        Yutong Yang$^{1,2}$,
        Zhaoxi Zhang$^{1,3}$,
        Sanwoo Lee$^{1,2}$,
        Yunfang Wu$^{1,2}$\thanks{~~Corresponding author.} \\
    $^{1}$National Key Laboratory for Multimedia Information Processing, Peking University \\ 
    $^{2}$School of Computer Science, Peking University, Beijing, China \\
    $^{3}$School of Computer Science \& Technology, Beijing Institute of Technology, Beijing, China \\
    \texttt{\{huang.hsiuyuan\}@stu.pku.edu.cn},
    \texttt{\{yytpku, sanwoo, wuyf\}@pku.edu.cn},
    \texttt{\{1120210536\}@bit.edu.cn}
    }

\usepackage{amsmath}
\usepackage{amsfonts}

\begin{document}
\maketitle
\begin{abstract}
As large language models (LLMs) continue to evolve, understanding and quantifying the uncertainty in their predictions is critical for enhancing application credibility. However, the existing literature relevant to LLM uncertainty estimation often relies on heuristic approaches, lacking systematic classification of the methods. In this survey, we clarify the definitions of uncertainty and confidence, highlighting their distinctions and implications for model predictions. On this basis, we integrate theoretical perspectives—including Bayesian inference, information theory, and ensemble strategies—to categorize various classes of uncertainty estimation methods derived from heuristic approaches. Additionally, we address challenges that arise when applying these methods to LLMs. We also explore techniques for incorporating uncertainty into diverse applications, including out-of-distribution detection, data annotation, and question clarification. Our review provides insights into uncertainty estimation from both definitional and theoretical angles, contributing to a comprehensive understanding of this critical aspect in LLMs. We aim to inspire the development of more reliable and effective uncertainty estimation approaches for LLMs in real-world scenarios.

% By systematically analyzing and synthesizing existing methods, w

% As large language models (LLMs) continue to proliferate, understanding and quantifying uncertainty in their predictions has become critical for enhancing trustworthiness in applications. 
% This survey aims to provide a comprehensive exploration of uncertainty estimation and application for LLMs, spanning multiple theoretical perspectives, including Bayesian inference, information theory and ensemble strategy. We investigate how these methods can address challenges specific to LLMs, such as hallucinations and overconfidence. Furthermore, we display approaches integrating uncertainty into diverse applications, including out-of-distribution detection, data annotation and question clarification. 
% Besides, we examine the distinctions between uncertainty and confidence, exploring how they are expressed in model outputs and their implications for decision-making. Our review highlights the importance of robust uncertainty quantification techniques in enabling LLMs to operate reliably in high-stakes environments, where transparency and accuracy are paramount. 

\end{abstract}

\section{Introduction}
As large language models (LLMs) continue to proliferate across various applications, understanding and quantifying their uncertainty has become increasingly important. 
% Uncertainty estimation作用1：provides valuable insights into the confidence of model predictions
Uncertainty estimation provides valuable insights into the confidence of model predictions, which is crucial for decision-making in high-stakes fields such as medical diagnosis \cite{fox1980evolution,simpkin2016tolerating},
% and autonomous systems \cite{6957722,9815528}, 
where incorrect predictions can have serious consequences \citep{alkaissi2023artificial, shen2023chatgpt}.

% Uncertainty estimation作用2：缓解LLM幻觉的手段
Moreover, uncertainty estimation can play a critical role in mitigating hallucinations in LLMs by providing an indication when answering questions outside their knowledge boundary \citep{li-etal-2023-halueval, huang2023survey, xu2024rejection}. Without effective measures of uncertainty in transformer-based systems, relying on generated language as a trustworthy source of information becomes difficult \cite{kuhn2023semantic}.

% LLM Uncertainty estimation 面临的困境
Despite the mature theoretical frameworks and practical applications for uncertainty estimation established in machine learning, these frameworks are typically model-specific \cite{banerjee2024llmshallucinateneedlive} or may not be adaptable to emerging LLMs due to two main factors. First, LLMs encompass an immense number of parameters, which makes the computational costs of traditional uncertainty estimation methods prohibitively high \cite{arteaga2024hallucinationdetectionllmsfast}. Second, the widespread use of commercial black-box API models complicates matters further, as these models often lack transparency and provide no access to internal parameters or output probabilities \cite{xiong2024llmsexpressuncertaintyempirical}. As a result, traditional uncertainty estimation approaches become impractical \cite{lin2024generatingconfidenceuncertaintyquantification}. 
% Furthermore, existing methods frequently struggle to quantify uncertainty on a continuous basis, limiting their effectiveness in delivering nuanced insights into model predictions. 
Therefore, there is an urgent need for a new, generalized, and reliable uncertainty estimation scheme tailored for LLMs.

% 现有综述的不足之处
%Most existing review articles provide a comprehensive overview of uncertainty estimation methods and cover some mainstream approaches 
Recently, there have been several review articles on uncertainty estimation methods \cite{gawlikowski2023survey, geng-etal-2024-survey}. However, many of these papers explore uncertainty estimation in a heuristic manner, focusing on issues like hallucination \cite{zhang2023sirenssongaiocean}, or may not clearly differentiate between confidence and uncertainty, which can lead to misunderstandings in the field.

% However, many of these papers explore uncertainty estimation in a heuristic manner, focusing on issues like hallucination \cite{zhang2023sirenssongaiocean}, or, more concerningly, they do not clearly distinguish between confidence and uncertainty, creating substantial confusion and misguiding the field.

% deviate significantly from the proper definitions of uncertainty. 
% For instance, concepts like \textbf{entropy}, which inherently describe uncertainty, are \textbf{incorrectly misclassified} as \textit{confidence estimation} methods by \citet{geng-etal-2024-survey}, creating substantial confusion and misguiding the field.
% However, many of these papers describe uncertainty estimation in a heuristic manner, sometimes straying from the theoretical foundations of uncertainty. 
We argue that grounding these methods in a clear theoretical framework is crucial for helping readers fully understand the concepts and inspiring future researchers to address the challenges mentioned above. Therefore, this review aims to bridge that gap by offering a more theory-driven exploration of uncertainty estimation methods.

\begin{figure*}[t!]
    \vspace{-1.0cm}
    \centering
    \resizebox{\textwidth}{!}{
        \begin{forest}
            forked edges,
            for tree={
                grow=east,
                reversed=true,
                anchor=base west,
                parent anchor=east,
                child anchor=west,
                base=left,
                font=\large,
                rectangle,
                draw=hidden-draw,
                rounded corners,
                align=left,
                minimum width=3em,
                edge+={darkgray, line width=1pt},
                s sep=3pt,
                inner xsep=2pt,
                inner ysep=3pt,
                line width=0.8pt,
                ver/.style={rotate=90, child anchor=north, parent anchor=south, anchor=center},
            },
            where level=1{text width=7.5em, font=\normalsize, align=center, text centered}{},
            where level=2{text width=15em, font=\normalsize, align=center, text centered}{},
            where level=3{text width=7.5em, font=\normalsize}{},
            where level=4{text width=6.0em, font=\normalsize}{},
            [
                Uncertainty, ver
                [
                    Uncertainty\\Estimation
                    [ 
                        UE with Bayesian Inference
                        [
                            BNNs (\S \ref{sec:AppBNN})
                            [
                                ~\cite{shridhar2019comprehensiveguidebayesianconvolutional,blundell2015weight},leaf,fill=grey!20, text width=42.0em
                            ]
                        ]
                        [
                            Approximate\\BNN Methods\\ (\S \ref{sec:AppBNN})
                            [
                                Variational Inference~\cite{graves2011practical,jordan1999introduction,1320776d-9e76-337e-a755-73010b6e4b64}\\Deep Gaussian Processes~\cite{iwata2017improvingoutputuncertaintyestimation,liu2020simpleprincipleduncertaintyestimation}\\Markov Chain Monte Carlo~\cite{xiao2018quantifyinguncertaintiesnaturallanguage}, leaf,fill=grey!20, text width=42.0em
                            ]
                        ]
                    ]    
                    [
                        UE with Ensemble Strategies, fill=green!10
                        [
                            Variance-\\Based Methods\\  (\S \ref{sec:variance}), fill=green!10
                            [MC Dropout~\cite{srivastava2014dropout,pmlr-v48-gal16,lakshminarayanan2017simple}\\Deep Ensemble~\cite{fadeeva2023lm,lakshminarayanan2017simple}\\BatchEnsemble~\cite{gal2016dropout,lakshminarayanan2017simple,wen2020batchensemblealternativeapproachefficient}\\\cite{,arteaga2024hallucinationdetectionllmsfast}, leaf,fill=grey!20, text width=42.0em
                            ]
                        ]
                        [
                            Consistency-\\Based Methods\\  (\S \ref{sec:consistency}), fill=green!10
                            [
                                Vanilla, fill=green!10
                                [
                                    ~\cite{selfconsistency2023,selectivelyanswering2023,huang2024uncttp}, leaf,fill=grey!20, text width=34.4em
                                ]
                            ]
                            [
                                Adversarial, fill=green!10
                                [\cite{zhang2024sac3reliablehallucinationdetection}\\\cite{lakshminarayanan2017simple,gawlikowski2023survey}\\\cite{huang2024uncttp},leaf,fill=grey!20,text width=34.4em
                                ]
                            ]
                            [
                                Re-Verified, fill=green!10
                                [~\citet{manakul2023selfcheckgpt,chen-mueller-2024-quantifying}, leaf,fill=grey!20, text width=34.4em   
                                ]
                            ]
                        ]
                        [
                            Similarity-\\Based Methods\\  (\S \ref{sec:similarity}), fill=green!10
                            [EigV~\cite{lin2024generatingconfidenceuncertaintyquantification}\\Degree matrix~\cite{lin2024generatingconfidenceuncertaintyquantification}\\Eccentricity~\cite{lin2024generatingconfidenceuncertaintyquantification}, leaf,fill=grey!20, text width=42.0em   
                            ]
                        ]
                    ] 
                    [
                        UE Based on Information Theory, fill=cyan!20
                        [
                            Entropy (\S \ref{sec:entropy}), fill=cyan!20
                            [    
                                ~\cite{kadavath2022language, kuhn2023semantic, duan2024gtbench},leaf,fill=grey!20, text width=42.0em
                            ]
                        ]
                        [
                            Perplexity (\S \ref{sec:perplexity}), fill=cyan!20
                            [
                                ~\cite{mora2024uncertainty,margatina2023active},leaf,fill=grey!20, text width=42.0em
                            ]
                        ]
                        [
                            Mutual\\Information (\S \ref{sec:mutualInformation}), fill=cyan!20
                            [
                                ~\cite{Ashinformation,malinin2019uncertainty,pmlr-v216-wimmer23a,Depeweg2019ModelingEA},leaf,fill=grey!20, text width=42.0em
                            ]
                        ]
                    ]
                    [
                        UE Using Language Expressions, fill=green!10
                        [
                            Verbalized \\Uncertainty (\S \ref{sec: Language}), fill=green!10
                            [
                                ~\cite{cosmides1996humans,lin2022teachingmodelsexpressuncertainty,tian2023justaskcalibrationstrategies}\\~\cite{xiong2024llmsexpressuncertaintyempirical,ZeroShot_CoT2022,groot2024overconfidencekeyverbalizeduncertainty},leaf,fill=grey!20, text width=42.0em
                            ]
                        ]
                    ]
                ]
                [
                    Uncertainty\\Application
                    [
                        Out-of-Distribution Detection (\S \ref{sec:ood}) 
                        [
                            ~\cite{LAMBERT2024102830,Mukhoti_2023_CVPR,srivastava2014dropout,NEURIPS2018_abdeb6f5,ren2022out},leaf,fill=grey!20, text width=51.14em
                        ]
                    ]
                    [
                        Data Annotation (\S \ref{sec:dataAnnotation})
                        [
                            Mitigate annotator-induced randomness~\cite{zhu2023investigating,Ge_Hu_Ma_Liu_Zhang_2024,he2024uncertaintyestimationsequentiallabeling,liu2022uamner}\\Active Data Annotation~\cite{cohn1996active,shen2017deep,schroder2021small}\\~\cite{margatina2021importance,diao2023active,margatina2023active},leaf,fill=grey!20, text width=51.14em
                        ]
                    ]
                    [
                        Question Clarification (\S \ref{sec:questionClarification})[~\cite{manggala2023aligning,hou2023decomposing,cole2023selectively,yona2024narrowing},leaf,fill=grey!20, text width=51.14em
                        ]
                    ]
                ]
            ]
        \end{forest}

    }
    \caption{Taxonomy of uncertainty estimates. Blue nodes stand for white-box-LLMs-only methods, while green nodes stand for methods suitable for black-box LLMs as well.}
    \label{taxo_of_icl}
\end{figure*}
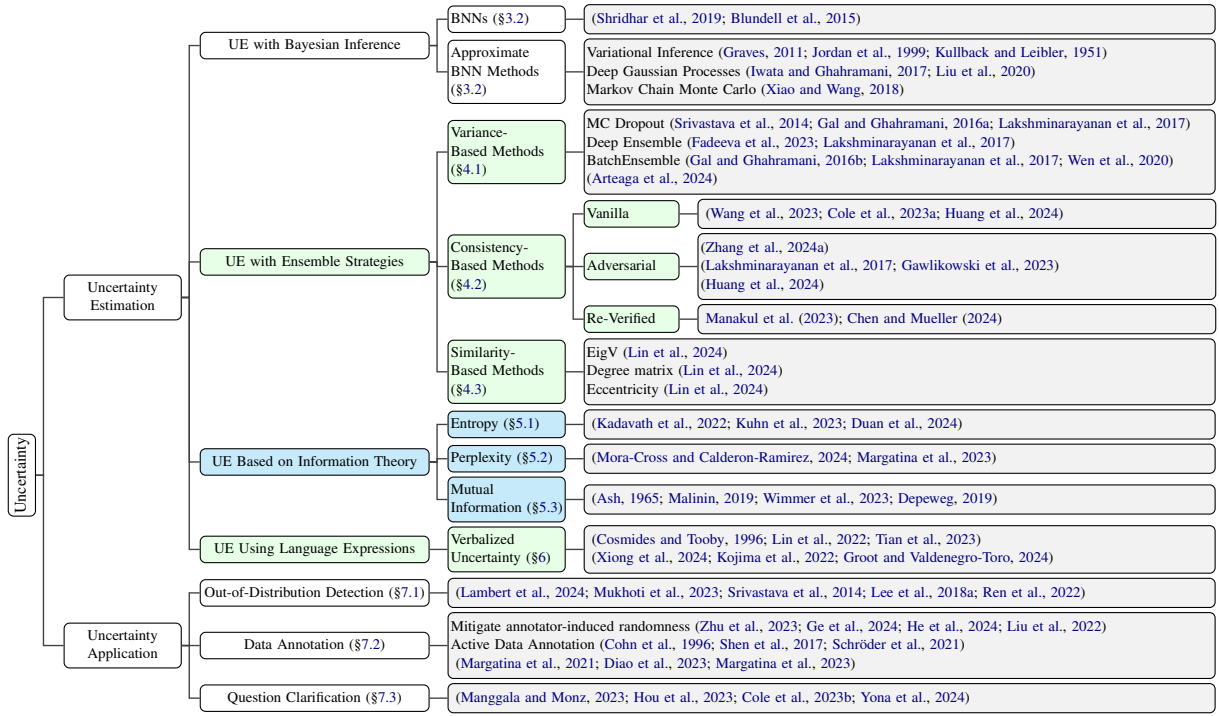

We begin by clarifying some easily confused concepts (Section \ref{sec:pre}). Next, we introduce the cornerstone of uncertainty estimation: Bayesian inference (Section \ref{sec: BP}). These methods rely on modeling the distributions of model parameters, which makes them not directly applicable to LLMs. However, it is still possible to indirectly incorporate Bayesian ideas for uncertainty estimation through various approximation techniques, often heuristic in nature. 
Following this, we discuss ensemble strategy (Section \ref{sec: Ensemble}), a non-Bayesian approach commonly used to approximate distributions, and we tailor this concept specifically for LLMs.
Then, we illustrate the uncertainty in LLMs through the lens of information theory (Section \ref{sec: Info}), using entropy,
%—a well-known concept that measures the degree of dispersion in a probability distribution—as well as its LLM-specific variant, 
perplexity, and mutual information.
Additionally, we explore existing verbal-based approaches to uncertainty estimation from the perspective of language expression (Section \ref{sec: Language}), a unique characteristic of LLMs.
These multi-faceted perspectives allow us to present a comprehensive understanding of uncertainty estimation in LLMs, bridging the gap between theoretical foundations and practical methods. In Section \ref{sec:application}, we demonstrate the task highly related to uncertainty. Figure \ref{taxo_of_icl} provides an overview of the article's structure.
% In Section \ref{sec:pre}, we begin by describing some important concepts such as the category of uncertainty, and clarify two easily confused concepts, uncertainty and confidence. We then delve into the definition of uncertainty and explore its estimation from four distinct perspectives. 
% These perspectives include Bayesian inference (Section \ref{sec: BP}), information theory (Section \ref{sec: Info}), ensemble strategy (Section \ref{sec: Ensemble}), and language expression (Section \ref{sec: Language}). 

% \begin{figure*}[!t]
%   \centering
%     \centering
%     \includegraphics[width=1\linewidth]{pics/abstract.pdf}
%     \caption{Place holder. Starting from the definition of uncertainty, we address LLM uncertainty from four theoretical perspectives}
%     %\caption{Overview of Article Structure}
%     \label{pic:abstract}
% \end{figure*}

\section{Preliminary}
\label{sec:pre}
%In this section, we present a preliminary of uncertainty estimation %from traditional machine learning to today's LLM 
%by clarifying some easily confused concepts.

\subsection{Uncertainty: Aleatoric vs. Epistemic}
According to \citet{AleatoryOrEpistemic}, there are two main types of uncertainty: epistemic (systematic) uncertainty that is caused by exceeding knowledge boundaries or lack of data, which can be reduced by expanding the training data; aleatoric (statistical) uncertainty that captures the inherent randomness within the experiment, which is inevitable in nature. 
% In other words, epistemic uncertainty refers to the reducible part of the (total) uncertainty, whereas aleatoric uncertainty refers to the irreducible part \cite{hullermeier_aleatoric_2021}. 

Although there is no consensus on whether these two types of uncertainty (epistemic and aleatoric) should be strictly separated in machine learning \cite{hullermeier_aleatoric_2021}, clarifying the distinction can help us better understand the challenges of uncertainty estimation in LLMs.
(1) Since the training data for LLMs is either unknown or too vast to retrieve realistically, confirming whether a specific piece of knowledge falls outside the model's learned scope is impractical, %evaluating epistemic uncertainty more difficult.
making the evaluation of epistemic uncertainty %more 
difficult.
(2) Additionally, the varying decoding strategies used in LLM generation complicate the evaluation of aleatoric uncertainty, as the noise inherent in the generation process is harder to quantify.

According to \citet{LAMBERT2024102830}, many existing uncertainty estimation works (61.95\%) assess total uncertainty rather than distinguishing between specific types. This paper primarily addresses total uncertainty, however, we also highlight the importance of differentiating between uncertainty types in specific contexts, such as question clarification (Section \ref{sec:questionClarification}). It is worth noticing that in the realm of LLM, addressing and mitigating epistemic uncertainty should be our ultimate goal. A reliable AI assistant should be able to recognize when a situation exceeds its knowledge boundaries and either prompt human intervention or refuse to provide answers.

% \subsection{Difference: Uncertainty estimates/estimation vs. uncertainty measures/quality}
% Uncertainty estimates refer to methods for generating or summarizing the uncertainty in neural network predictions \cite{gawlikowski2023survey}, focusing on identifying sources like model and data uncertainty to assess reliability. In contrast, uncertainty measures are the metrics used to evaluate the quality and trustworthiness of these estimates \cite{gawlikowski2023survey}. Evaluating uncertainty is challenging due to the lack of ground truth values and a unified metric across tasks.

% In other words, uncertainty estimates provide methods to derive uncertainty values, while uncertainty measures assess the appropriateness of these methods \cite{gawlikowski2023survey}. This article focuses on uncertainty estimates (uncertainty estimation).

\subsection{Differences: Uncertainty vs. Confidence}
Uncertainty and confidence are distinct yet interrelated aspects of model evaluation, particularly in LLMs. While some researchers suggest that increased uncertainty correlates with decreased confidence \cite{geng-etal-2024-survey,xiao-etal-2022-uncertainty,chen2023quantifyinguncertaintyanswerslanguage}, this view lacks a clear distinction between the two concepts. Following \citet{lin2024generatingconfidenceuncertaintyquantification}, for \(P(Y|x)=\mathcal{N}(\mu, \sigma^2)\), \(\sigma^2\) represents uncertainty, while confidence in output \(Y = y_0\) is expressed as \(-\frac{y_0-\mu}{\sigma}\). For instance, in a classification task, low uncertainty signifies a dominant class probability, which correlates with high confidence. However, high confidence does not necessarily imply low uncertainty, as the probabilities of other classes contribute to the overall uncertainty.

In brief, uncertainty refers to the overall output distribution, indicating the variability in potential predictions, whereas confidence pertains to a specific prediction, denoting the likelihood of that output \cite{manakul2023selfcheckgpt}. The existing works mainly focus on aggregating \textbf{multiple responses} to get the most accurate answer, as shown in Section \ref{sec: Ensemble}, than focus on one particular prediction. Therefore, we argue that “uncertainty” is a more precise term than “confidence” in the context of estimation. In this article, we emphasize uncertainty estimation.

\section{Uncertainty Estimation with Bayesian Inference}
\label{sec: BP}
%In this section, we examine the problem of uncertainty estimation from a Bayesian perspective. We begin by outlining the principles of Bayesian theory and its practical implementation. 
The Bayesian 
%perspective 
theory
is crucial for estimating uncertainty, as most methodologies draw upon concepts derived from Bayesian theory to varying degrees. 

\subsection{Bayesian Neural Networks}
\label{sec:BNNs}
% 修改一下：
% BNN将模型权重视为概率分布来捕捉模型对数据的信念。在**训练**过程中，BNN利用先验知识和观测数据来更新权重的后验分布，这个后验分布反映了模型在给定数据后的不确定性。具体来说，后验分布的宽度和形状可以揭示模型对预测的信心程度：较宽的后验分布表示较大的不确定性，而较窄的后验分布则表明模型对预测结果更有信心。因此，BNN通过提供一个关于模型权重的不确定性描述，使得对输入数据的预测不仅依赖于单一值，而是考虑了权重的不确定性，从而更全面地评估模型的预测可信度。

Bayesian Neural Networks (BNNs) estimate the posterior distribution of weights $p(w|\mathcal{D})$ based on training data D, which is crucial in Bayesian inference as it represents the updating of beliefs about model weights with increasing data \cite{shridhar2019comprehensiveguidebayesianconvolutional}. By taking the expectation over the posterior, BNNs calculate the predictive distribution of label $y$ given input $x$ as $\mathbb{E}_{p(w|D)} \left[  P(y|x, w) \right]$, effectively incorporating uncertainty into their predictions. In contrast to traditional Deep Neural Networks (DNNs) that rely on point estimates, typically optimized via maximum likelihood or maximum a posterior estimation, BNNs provide a more comprehensive view of model uncertainty.

\subsection{Approximate BNN Methods}
\label{sec:AppBNN}
% 修改一下：
% **推理**时，我们从这些**分布中抽样（采样）不同的权重进行多次预测**。通过观察多次预测结果的差异，我们可以得到模型的不确定性估计。要真正实现贝叶斯神经网络，我们需要**对每个模型参数进行采样**，并**计算其后验分布**。然而，由于实际问题往往非常复杂，这个过程**计算代价高昂**，通常通过一些**近似**的方法来实现。

While BNNs provide a principled framework for uncertainty estimation, 
% by computing the posterior distribution of model weights through sampling
the complexity and computational cost of obtaining the exact posterior is often unacceptable. To address this, approximate BNN methods have been developed to efficiently approximate the posterior distribution.
% , aiming to strike a balance between computational efficiency and maintaining the robustness of uncertainty estimates, thereby making Bayesian approaches more accessible and applicable in diverse scenarios.

\paragraph{Variational Inference}
% Variational Inference (VI) is a crucial approach for estimating model uncertainty \cite{graves2011practical}. Since the posterior distribution cannot be computed analytically, a simpler variational distribution is constructed to approximate the true posterior \cite{jordan1999introduction}.
Variational Inference (VI) is a crucial approach for estimating model uncertainty \cite{graves2011practical}, which involves constructing a simpler variational distribution to approximate the true posterior \cite{jordan1999introduction}. Theoretically, the Kullback-Leibler (KL) divergence \cite{1320776d-9e76-337e-a755-73010b6e4b64} measures the difference between these two distributions. However, computing the KL divergence directly is often intractable. 
Instead, the Evidence Lower Bound (ELBO) \cite{jordan1999introduction} is more commonly employed to approximate the posterior. 
% In BNNs, multiple outputs are obtained by sampling from the weight distribution, and the ELBO is used to optimize the parameters of the variational distribution \cite{graves2011practical}.

% \small \begin{equation}
%     \log P(D) \geq \mathbb{E}_{q(\theta)}[\log P(D | \theta)] - \text{KL}(q(\theta) \| P(\theta))
% \end{equation} \normalsize

\paragraph{Deep Gaussian Processes }
Deep Gaussian Processes (DGP) are flexible, non-parametric models that leverage Bayesian inference to predict outcomes by estimating posterior distributions from prior data \cite{iwata2017improvingoutputuncertaintyestimation}. \citet{liu2020simpleprincipleduncertaintyestimation} introduces the concept of distance awareness, which measures the similarity between inference samples and training data. While DGPs are adaptable and perform well with small datasets, they come with high computational costs and scalability challenges. Moreover, the choice of kernel function is critical and requires careful tuning to optimize performance.
DGPs are adaptable and perform well with small datasets but are computationally expensive and challenging to scale. The choice of kernel function is critical and requires tuning experience.
% DGPs are adaptable and perform well with small datasets but are computationally expensive and challenging to scale. The choice of kernel function is critical and requires tuning experience.

\paragraph{Markov Chain Monte Carlo}
Markov Chain Monte Carlo (MCMC) generates samples that approximate posterior distribution by constructing a sequence of states in a Markov chain. Each new state relies solely on the current state, ensuring that the process retains the Markov property. As the chain progresses, the samples converge to the desired posterior distribution. \citet{xiao2018quantifyinguncertaintiesnaturallanguage} applies MCMC in the context of semantic segmentation, demonstrating its utility.

\subsection{Takeaways}
BNNs' inference provides both mean and variance, which are helpful in tasks needing high-confidence predictions like autonomous driving and medical diagnosis. Although BNNs offer a flexible framework for defining priors, selecting an appropriate prior for specific tasks remains challenging. Noise Contrastive Priors (NCP) \cite{hafner2019noisecontrastivepriorsfunctional} proposes a method by adding noise to input data using a prior, and a wide distribution as the output prior. Considering the resource requirements, variational inference is the most ubiquitous method. 

\section{Uncertainty Estimation with Ensemble Strategies}
%Ensemble Perspective of Uncertainty}
\label{sec: Ensemble}

% \begin{figure}[!t]
%   \centering
%     \centering
%     \includegraphics[width=1\linewidth]{pics/ensemble.png}
%     \caption{Place holder}
%     \label{pic:ensemble}
% \end{figure}

Ensemble methods offer a robust framework to enhance predictive performance and, more pertinent to our focus, to provide uncertainty estimations. 
In this section, we summarize three mainstream approaches for integrating model outputs to assess uncertainty, 
%—specifically, by evaluating response dispersion.
% —as illustrated in Figure \ref{pic:ensemble}. 
%These approaches
including variance-based, consistency-based and similarity-based ensemble techniques.
%It's important to note that while the discussions of ensembles in this section are framed from a Bayesian perspective, we utilize a variety of non-Bayesian methods to construct these ensembles \cite{malinin2019uncertainty}.

% In this section, we summarize three mainstream approaches for integrating model outputs to assess uncertainty, as shown in Figure \ref{pic:ensemble}: variance-based, consistency-based, and similarity-based ensemble techniques. It's important to note that while the discussions of ensembles in this section are framed from a Bayesian perspective, we utilize a variety of non-Bayesian methods to construct these ensembles \cite{malinin2019uncertainty}.

% In this section, we will study uncertainty through the lens of a high-level concept abstracted from ensemble learning, as shown in \ref{pic:ensemble}.
%We will examine several uncertainty measures that are (theoretically) applicable to LLMs from an ensemble perspective. Discussions over ensemble-based methods for traditional DNNs will be in the appendix. 

% It is necessary to point out that while the discussion of ensembles so far has been from a Bayesian viewpoint, it is possible to construct ensembles using a range of non-Bayesian approaches \cite{malinin2019uncertainty}.

% \subsection{White-Box Uncertainty Estimation}
\subsection{Variance-Based Ensemble}
\label{sec:variance}
Ensemble-based methods for DNNs measure uncertainty by leveraging the diversity of predictions generated by multiple model versions under slightly varied conditions \cite{fadeeva2023lm}. Specifically, the \textit{variance} between the predictions of the individual predictor can be seen as a natural vehicle for uncertainty estimation, with higher disagreement indicating higher uncertainty. The final prediction is calculated by averaging each model's output \cite{lakshminarayanan2017simple}, using:

\small \begin{equation}p(y|\mathbf{x}) = M^{-1} \sum_{m=1}^{M} p_{\theta_m} \left( y|\mathbf{x}, \theta_m \right)
\end{equation} \normalsize
where \( M\) indicates the set of models, and the variance can be calculated as : 

\small \begin{equation}\sigma_{*}^{2}(\mathbf{x}) =M^{-1} \sum_{m} \left( \sigma_{\theta_m}^{2}(\mathbf{x}) + \mu_{\theta_m}^{2}(\mathbf{x}) \right) -\\ \mu_{*}^{2}(\mathbf{x})
\end{equation} \normalsize

% However, as the parameters scale to LLMs, modifying multiple LLMs becomes prohibitively expensive. In this section, we will introduce two classical white-box DNN uncertainty estimation methods—Monte Carlo Dropout \cite{gal2016dropout} and Deep Ensembles \cite{lakshminarayanan2017simple}—that are primarily theoretical in their applicability to LLMs. Our goal is to inspire the development of future methods that can be effectively applied to white-box LLMs.

\paragraph{Monte Carlo Dropout}
Dropout \cite{srivastava2014dropout} is a widely used regularization technique in neural networks, designed to prevent overfitting during the training phase by randomly discarding a fraction of neurons. Monte Carlo Dropout (MCD) \cite{pmlr-v48-gal16} extends this concept to the inference phase. By retaining the dropout mechanism during inference and randomly dropping neurons at each iteration, MCD generates multiple output samples. These samples can then be used to estimate the variance of the predictive distribution, providing valuable insights into the model's uncertainty regarding its predictions.

% Monte Carlo Dropout (MCD) samples network outputs using the idea of dropout \cite{srivastava2014dropout}. With dropout, network parameters have a probability of being frozen during forward and backward passes. MCD extends this to inference \cite{pmlr-v48-gal16}, providing a probabilistic interpretation of dropout and introducing mathematical support to estimate uncertainty from a trained DNN. During inference, the network randomly zeros parameters, generating a distribution of outputs from which uncertainty and other key information can be derived.

\paragraph{Deep Ensemble}

Deep Ensemble, introduced by \citet{lakshminarayanan2017simple}, has emerged as a state-of-the-art method for uncertainty estimation. Unlike traditional single-model approaches, Deep Ensemble leverages multiple models trained independently on either randomly sampled subsets or the entire dataset \cite{lakshminarayanan2017simple}.

\paragraph{Batch Ensemble}
Batch Ensemble \cite{wen2020batchensemblealternativeapproachefficient}, a more lightweight and parallelizable variant of deep ensemble, introduces the concepts of “slow weights” and “fast weights.” The slow weights serve as the base weights, while the fast weights provide modifications to these base weights, allowing Batch Ensemble to be effectively employed with LLMs \cite{arteaga2024hallucinationdetectionllmsfast}.

\subsection{Consistency-Based Ensemble}
\label{sec:consistency}
Research has shown that higher predictive uncertainty is associated with an increased likelihood of hallucinations \cite{xiao2021hallucinationpredictiveuncertaintyconditional}. 
Uncertainty estimation methods for black-box LLMs largely rely on heuristic approaches to mitigate hallucinations, linking uncertainty to both \textit{confidence} and \textit{hallucination scores} \cite{zhang-etal-2023-sac3, manakul2023selfcheckgpt}.
We propose framing these heuristic methods within uncertainty estimation approaches from an ensemble perspective by focusing on how to evaluate response consistency, without strictly distinguishing between these concepts.

% We have further summarized three main categories for how to determine consistency between answers, which is \textit{vanilla}, \textit{adversarial}, and \textit{re-verified}. Although they come in various forms and it is not conclusive which one is the best, each provides a unique perspective on uncertainty. 

\paragraph{Vanilla Methods}

Inspired by \citet{chen-mueller-2024-quantifying}, we categorize methods for deriving uncertainty estimates from \textbf{user observations} of LLM responses as vanilla methods.
The idea that answer consistency reflects uncertainty in LLMs is supported by \citet{selfconsistency2023}, who propose self-consistency (or temperature sampling) and find that consistency correlates highly with accuracy. This suggests that low consistency indicates uncertain and therefore confers some ability for the model to “know when it doesn't know”. 
\citet{selectivelyanswering2023} introduced two metrics—\textit{repetition} and \textit{diversity}—to measure consistency. Let $O$ denote the outputs of the LLM, $O_g$ represent the greedy output, and $S$ be the set of sampled outputs. Repetition is expressed as:

\small \begin{equation}
Repetition = \frac{|\{o \in S \mid o = O_g\}|}{|S|}
\end{equation} \normalsize
Diversity, inversely proportional to the number of distinct samples, is defined by:

\small \begin{equation}
Diversity = 1 - \frac{|S|}{|Distinct(S)|}
\end{equation} \normalsize
Here, $|Distinct(S)|$ is the count of unique samples in $S$; a value of zero is assigned if all samples are distinct. Additionally, \citet{huang2024uncttp} framed uncertainty quantification in LLMs as a binary problem, where the LLM inconsistency is considered to be \textit{uncertain} and vice versa \textit{certain} after a definite number of samplings.

% Repetition is calculated as the fraction of times the sampled outputs match the greedy output, while diversity is inversely proportional to the number of distinct samples, with a value of zero assigned if all samples are different.

% \citet{huang2024uncttp} intuitively treated the quantification of uncertainty in the LLM as a binary problem, where the LLM inconsistency is considered to be \textit{uncertain} and vice versa \textit{certain} after a definite number of samplings.

\paragraph{Adversarial Methods}
% One of the most crucial points when applying ensemble methods is to maximize the variety in the behaviour among the single networks \cite{lakshminarayanan2017simple,RENDA20191, gawlikowski2023survey}. The adversarial methods we summarised build upon vanilla approaches by adding a twist of adversarial components, which introduces a larger variety among the different responses. 
Maximizing diversity among individual networks is crucial when applying ensemble methods \cite{lakshminarayanan2017simple,RENDA20191, gawlikowski2023survey}. The adversarial methods we summarised build upon vanilla approaches by adding a twist of adversarial components, which increase variability in LLM responses.
\citet{zhang2024sac3reliablehallucinationdetection} introduce a mechanism that perturbs semantically equivalent questions to evaluate the consistency of LLM responses across variations of the same question.
Additionally, \citet{huang2024uncttp} inject correct and incorrect labels, respectively, into the prompt during sampling, in addition to using the vanilla method. The uncertainty level is then determined by the LLM responses' consistency across three samplings for each instance. If the results are consistent, the model is classified as \textit{certain}; otherwise \textit{uncertain}.

% \citet{kadavath2022languagemodelsmostlyknow} discover injecting random answer into the prompt can help improve the inference performance. Yet it dose not explicitly define it as uncertainty estimation.

\paragraph{Re-Verified Methods}
% 本章我们凭什么认为Hallucination Score==Uncertainty Score -------
% SelfCheckGPT: Zero-Resource Black-Box Hallucination Detection for Generative Large Language Models
% 然而，对于不同的输入，例如“John Smith is a _”，系统将不确定连续性，这可能会导致平坦的概率分布。在推理过程中，这可能会导致生成非事实词。这种洞察力使我们能够理解不确定性指标与事实性之间的联系。事实句子可能包含具有较高似然性和较低熵的标记，而幻觉可能来自具有高度不确定性的平坦概率分布的位置。
% -------------------------------------------------------------
% 方法
The biggest difference between re-verified and other consistency-based methods lies in the round of interactions. These methods check the LLM's consistency by having the LLM itself, or another model, answer a closed-ended fact-checking question. While they are typically designed to mitigate LLM hallucinations, they can also be regarded as a means of evaluating consistency.
\citet{manakul2023selfcheckgpt, chen-mueller-2024-quantifying} instructed the LLM to evaluate its response by selecting from a limited set of options, such as A) Correct, B) Incorrect, or C) Unsure, and then a numerical score is assigned to each option, and the average score across multiple rounds of such verification questions is computed to determine the LLM's uncertainty for each instance.

\subsection{Similarity-Based Ensemble}
\label{sec:similarity}
% These methods is proposed by \citet{lin2024generatingconfidenceuncertaintyquantification}, which calculates the similarity between multiple responses to indirectly quantify the dispersion of the model outputs. Compared to consistency-based methods, similarity-based methods can offer a more continuous measurement of uncertainty. Because the similarity is derived from the predicted probabilities of an off-the-shelf NLI model instead of assigning values to several variables and then averaging them.
These methods, proposed by \citet{lin2024generatingconfidenceuncertaintyquantification}, calculate the similarity between multiple responses to indirectly quantify the dispersion of model outputs. Compared to consistency-based methods, similarity-based approaches offer a more continuous measurement of uncertainty, as the similarity is derived from the predicted probabilities of an off-the-shelf Natural Language Inference (NLI) model rather than by assigning values to different variables and averaging them.
\citet{lin2024generatingconfidenceuncertaintyquantification} used a small NLI model to divide the semantic set of responses, where the predicted probabilities are viewed as the similarity. With the similarity between each answer obtained by the NLI model, we can construct the adjacency matrix. Then, the sum of eigenvalues of the graph Laplacian (EigV), degree matrix, and eccentricity can be calculated. For more detail, please refer to \citet{lin2024generatingconfidenceuncertaintyquantification}.

% \subsubsection{Reflective-based Methods}
% These methods involve directly querying the model to ascertain the veracity of a given answer.This approach aligns with self-verification-based methods(see \ref{chap non-bayesian methods}) commonly employed for black-box models; however, white-box models can leverage their internal probability distributions to make such judgments more directly.\\
% \citet{kadavath2022language} demonstrated that large language models exhibit good calibration when responding to True/False questions with human-authored samples. They introduced the \textbf{P(True)} method, which queries the model about its own generated samples by asking: "Is the proposed answer:(A)True (B)False"
% and subsequently evaluates the model's assigned probabilities for the ‘A’ and ‘B’ choices.\\
% Nonetheless, evaluating the model’s own outputs presents greater challenges, as these samples may reside near the model's decision boundary for determining validity, or the model might exhibit overconfidence in its own responses. To enhance performance, the model can be presented with additional samples labeled as T = 1 for comparative purposes. The study also found that larger models tend to perform better in such evaluative tasks.

\subsection{Takeaways}
Monte Carlo Dropout \cite{gal2016dropout} and Deep Ensembles \cite{lakshminarayanan2017simple} are two classical white-box variance-based ensemble techniques that efficiently approximate Bayesian inference for uncertainty estimation. However, modifying multiple models can become prohibitively expensive \cite{lakshminarayanan2017simple}, especially as the number of parameters scales up with LLMs, which renders these techniques primarily applicable in theory rather than practice for LLMs. In contrast, Batch Ensemble provides a practical solution to address this challenge.

As for the consistency or similarity based ensemble methods, they estimate uncertainty by evaluating the consistency or similarity of responses across multiple samples. While relatively straightforward, these methods come in various forms, each offering unique perspectives on uncertainty.

\section{Uncertainty Estimation Based on Information Theory}
\label{sec: Info}
% \subsection{White-Box LLM}
%这里待改，应该先总结一下Traditional methods，指出尽管对于BNN和DNN已经有一套完善的方法和框架，但是LLM的UE面临着一些独有的困难（LLM的输出是一个sequence，怎么计算一个sequence的不确定性？对于essemble方法：LLM参数量太大，有计算量过大的问题？对于density方法：没有训练数据的的问题）根据后面的内容修改。
%简述最后采取的分类方式
% 
% Following the discussion on deep learning models, we now turn to white-box LLM uncertainty estimation methods.
% These approaches require direct access to internal model information, such as logits, internal layer outputs, or the model itself.
%上面不用改了，现在是从信息论的角度来讲了，不是从模型的角度

Shannon established the foundations of information theory by integrating the principles of probability theory with the quantification of information, significantly impacting various fields, including machine learning. This section explores uncertainty estimation in LLMs from the information theory point of view.

%这部分现在融合了 Non-Bayesian method的部分
% Since BNNs replace deterministic weights with distributions, they inherently provide a mechanism for uncertainty estimation grounded in Bayesian theory. However, despite this theoretical advantage, BNNs have not achieved the same widespread adoption as DNNs and have yet to be successfully implemented in LLMs. 
%Consequently, the need to estimate uncertainty in DNNs remains critical. 
% In this section, we discuss uncertainty estimation methods based on information theory for LLMs.
%这里直接说基于信息论，待考虑对不对
% uncertainty estimation in DNNs. Although these approaches were initially developed for standard DNNs, they are applicable to LLMs.
%, as will be elaborated upon in the subsequent section. 

\subsection{Entropy}
\label{sec:entropy}
% In traditional classification tasks, entropy is a common metric for assessing the sharpness of output distributions, offering insights into uncertainty \cite{wang2022extractingguessingimprovingfaithfulness,malinin2018predictiveuncertaintyestimationprior}. 
% However, for LLMs applied to text generation, quantifying uncertainty across an entire sequence is more complex. 
For classification tasks, entropy indicates the degree of dispersion in the distribution of the model's predictions for a given input \cite{wang2022extractingguessingimprovingfaithfulness,malinin2018predictiveuncertaintyestimationprior}, which can be present in:

\small \begin{equation}
H(X) = - \sum_{i=1}^{n} p(x_i) \log p(x_i)
\end{equation} \normalsize
where $p_i$ is the model's prediction probability for category $i$. The higher the entropy value, the more uncertain the model's prediction. Quantifying uncertainty across an entire sequence in text generation tasks using LLMs is more complex. \citet{kadavath2022language} use the probability of the complete sequence to compute Predictive Entropy (PE). 
Given an input \( x \) and a generated sentence \( s \) consisting of \( N \) tokens, where \( s \) is a completion based on \( x \), the probability of generating the \( i \)-th token \( s_i \) given the preceding tokens \( s_{<i} \) and the prompt \( x \) is denoted as \( p(z_i \mid s_{<i}, x) \). The \textbf{PE} of the entire sentence \( s \) is given by:

\small \begin{equation}
    \textbf{PE}(\boldsymbol{s}, \boldsymbol{x}) = - \log p(\boldsymbol{s}|\boldsymbol{x}) = \sum_{i}-\log p(s_i|\boldsymbol{s}_{<i},\boldsymbol{x})
\end{equation} \normalsize
This token-level measure is commonly used as a baseline for assessing uncertainty.

However, free-form text generation presents unique challenges due to semantic equivalence, where different sentences can convey the same meaning. This can lead to inflated uncertainty estimates at the token level because different tokens might represent similar meanings. To address this, \textbf{Semantic Entropy (SE)} has been proposed \cite{kuhn2023semantic}. SE enhances token-level measures by clustering sentences into equivalence classes based on their semantic similarity and computing entropy over these classes:

\small \begin{equation}
SE(x) = - \sum_{c} p(c|x) \log p(c|x)
\end{equation} \normalsize
where \( c \) denotes an equivalence class of semantically similar sentences.

Recent work by \citet{duan2024gtbench} highlights that not all tokens contribute equally to the underlying meaning, as linguistic redundancy often allows a few key tokens to capture the essence of longer sentences. To improve uncertainty estimation, \citet{duan2024gtbench} introduce \textbf{SAR} (Shifting Attention to Relevance), a heuristic method that adjusts attention to more relevant components at both the token and sequence levels.

%这部分待放到关于metric的部分
% These methods assume that a model's confidence in its predictions is correlated with their accuracy; thus, higher uncertainty generally suggests a greater likelihood of an incorrect response. Consequently, \textbf{AUROC} (Area Under the Receiver Operating Characteristic Curve) is used as the evaluation metric. AUROC measures the probability that a randomly chosen correct response has a higher uncertainty score than a randomly chosen incorrect response, with a perfect uncertainty measure scoring 1 and a random measure scoring 0.5.

\subsection{Perplexity}
\label{sec:perplexity}
% Perplexity是度量语言模型的性能常见的指标之一，通过对平均交叉熵损失进行指数化计算得出。它反映了模型在生成下一个词时的“困惑”程度。如果模型分配的概率分布很分散，Perplexity 值会较高，意味着模型对下一个词的不确定性较大。
Perplexity is a widely used metric for evaluating the readability and accuracy of text generated by LLMs, which is calculated by exponentiating the average cross-entropy loss. It can be understood as a measure of how surprised the model is when evaluating a sequence of tokens \cite{mora2024uncertainty}. A higher perplexity value indicates that the model assigns a more dispersed probability distribution, signifying greater uncertainty about the next word. Therefore, perplexity can also serve as an indicator of LLM uncertainty.
% Perplexity is one of the common metrics for evaluating the performance of language models, obtained by exponentiating the average cross-entropy loss. It can be understood as a measure of how surprised the model is when evaluating a sequence of tokens. Therefore, perplexity can also serve as an indicator of uncertainty \cite{mora2024uncertainty}.
% The perplexity of generated text is computed using the following formula:
% \small \begin{equation} 
%     {PPL}(S) = \exp\bigg(-\frac1I \sum_{i=1}^I \log p(s_i|s_{<i})\bigg)
% \end{equation} \normalsize
% where $I$ is the length of the sequence, and $p(s_i|s_{<i})$ denotes the model's predicted probability for the token given the preceding tokens $s_{<i}$.
\citet{margatina2023active} uses perplexity to evaluate the uncertainty of each in-context example and select those with the highest perplexity as few-shot data for in-context learning (ICL).

\subsection{Mutual Information}
\label{sec:mutualInformation}
A basic result from information theory shows that Shannon entropy can be additively decomposed into conditional entropy and mutual information \cite{Ashinformation}: 

\small \begin{equation}
H(Y)=H(Y|\Theta)+I(Y,\Theta)
\end{equation} \normalsize

Conditional entropy \(H(Y|\Theta)\) represents the uncertainty remaining in Y when the realization of \(\Theta\) is known, and it naturally serves as a measure of aleatoric uncertainty. Thus, mutual information, which captures the difference between total uncertainty and aleatoric uncertainty, serves as a measure of epistemic uncertainty \cite{Depeweg2019ModelingEA}. 

\citet{malinin2019uncertainty} posits that mutual information measures the 'disagreement' between models in an ensemble, and therefore reflects epistemic uncertainty, which arises from the model’s lack of understanding of the data. However, \citet{pmlr-v216-wimmer23a} argue that mutual information is better interpreted as a measure of divergence or conflict rather than ignorance, may not be the right measure of epistemic uncertainty. 

\subsection{Takeaways}
% \paragraph{Black-Box Limitations}
Notably, information-based methods require access to token-level probabilities from LLMs, which makes them unsuitable for current black-box LLMs or API models. A potential solution is to utilize white-box LLMs as surrogate models to provide these probabilities. \citet{shrivastava2023llamas} demonstrates that probabilities from weaker white-box surrogate models can effectively estimate the internal confidence levels of stronger black-box models, such as GPT-4, and outperform linguistic uncertainty measures.

\section{Uncertainty Estimation Using Language Expressions}
\label{sec: Language}
%Next-token prediction endows LLMs with a unique capability to generate language, setting them apart from other deep neural networks. 
By training on vast amounts of natural language data, current LLMs are able to produce language that closely approximates human speech. %while also inheriting certain characteristics of human language to varying degrees. However, this capability comes with both strengths and limitations. In this section, we will examine uncertainty in LLMs from a linguistic perspective and explore its quantitative aspects.
% \subsection{Preliminary}
%\subsection{Verbalized Uncertainty}
% \subsection{Naturalistic Expression of Uncertainty}
A key aspect of human intelligence lies in our capability to express and communicate our uncertainty in a variety of ways \cite{cosmides1996humans}. Unlike uncertainty in the statistical sense, which relates to the degree of dispersion in output, this section examines LLM's uncertainty through the lens of naturalistic expressions. 
%without making a strict distinction between the concepts of confidence and uncertainty. 

This scope of research focuses on prompting LLMs to explicitly articulate their level of uncertainty alongside their responses. \citet{lin2022teachingmodelsexpressuncertainty} introduces the concept of verbalized confidence that prompts LLMs to express its uncertainty using natural language for representing degrees. 
% They mainly focus on fine-tuning on specific datasets where the confidence is provided. 
\citet{tian2023justaskcalibrationstrategies} proposes prompting LLMs to generate the top-k guesses and their corresponding confidence for a given question.
\citet{xiong2024llmsexpressuncertaintyempirical} proposes Self-Probing which is to ask LLMs “How likely is the above answer to be correct?” and have them verbalize its uncertainty in the form of the numerical number in the range of 0-100\%.

% \paragraph{Takeaways}
It is important to note that the conclusions regarding uncertainty estimation in this section vary by methods, tasks, evaluation metrics, and models, and these findings should be interpreted within their specific context. \citet{xiong2024llmsexpressuncertaintyempirical} conducts an empirical assessment of zero-shot verbal-based methods across different sampling and aggregation strategies, revealing that LLMs often exhibit overconfidence. This overconfidence can be mitigated by employing prompting strategies, such as zero-shot Chain of Thought (CoT) \cite{ZeroShot_CoT2022}. However, \citet{tian2023justaskcalibrationstrategies} reaches an opposite conclusion, indicating that CoT does not enhance verbalized calibration. Additionally, research has shown that the accuracy of verbal-based uncertainty estimation varies by task. For instance, in sentiment analysis, models tend to be underconfident, while overconfidence is observed in tasks such as math word problems and named entity recognition \cite{groot2024overconfidencekeyverbalizeduncertainty}.

\section{Uncertainty Application}
\label{sec:application}

\subsection{Out-of-Distribution Detection}
\label{sec:ood}
Out-of-distribution (OOD) data refers to samples that significantly deviate from the training data of a machine learning model \cite{LAMBERT2024102830}. In other words, when the model encounters inputs beyond its training knowledge, it typically exhibits high uncertainty.
% Epistemic uncertainty和OOD相辅相成
Epistemic uncertainty, which reflects the model's lack of knowledge about certain inputs, is interrelated with and complementary to OOD detection. On the one hand, by leveraging epistemic uncertainty, the accuracy of OOD detection can be enhanced, as high epistemic uncertainty often indicates that the sample is likely to be OOD \cite{Mukhoti_2023_CVPR}. On the other hand, understanding OOD data can help refine uncertainty estimation methods. By recognizing the high uncertainty models exhibited when faced with OOD data, researchers have developed techniques like Monte Carlo Dropout \cite{srivastava2014dropout}, significantly improving OOD detection.

% The uncertainty of DNNs is closely related to the problem of detecting Out-of-Distribution (OOD) samples. If the uncertainty of a sample exceeds a threshold, it can be considered as OOD. Intuitively, high uncertainty indicates a high likelihood of OOD. 
% \citet{Mukhoti_2023_CVPR} proposed Deep Deterministic Uncertainty to detect OOD samples based on leveraging epistemic uncertainty. After training the model on with a well-regularized feature space, the Gaussian Discriminant Analysis (GDA) is applied post-training to estimate the feature space density.
% At test time, if the feature-space density of a sample is low, this indicates high epistemic uncertainty, meaning the sample is likely OOD. 

%修改在下面一段，这里没动
Mahalanobis Distance (MD) is widely employed for OOD detection. Given a mean vector $\mu$ and a covariance matrix $\Sigma$, the MD is defined as:

\small \begin{equation}
MD(z_{test}; \Sigma, \mu) = (z_{test} - \mu)^T \Sigma^{-1} (z_{test} - \mu)
\end{equation} \normalsize

The method \cite{NEURIPS2018_abdeb6f5} calculates uncertainty using MD based on the distance to the nearest class-conditional Gaussian distribution. Greater deviation indicates higher uncertainty, suggesting the sample is likely OOD. 
%Input pre-processing and feature ensemble techniques further amplify uncertainty for better detection. This approach excels in detecting both OOD samples and adversarial attacks, demonstrating the importance of uncertainty in effective OOD detection.
%增加与density-based UE method的讨论
Notably, MD can also be considered a density-based uncertainty estimation method\cite{fadeeva2023lm}. Both OOD detection and density-based uncertainty measurement aim to identify data points that differ from the training data. In OOD detection, these points are treated as outliers, whereas in density-based approaches, they are often seen as residing in low-density regions, which correlate with higher uncertainty.

Based on MD, \citet{ren2022out} introduced the Relative Mahalanobis Distance (RMD), defined as:

\small \begin{equation}
MD(z_{test}) := MD(z_{test}; \mu^z, \Sigma^z)
\end{equation} \normalsize
\small \begin{equation}
RMD(z_{test}) :=\\ MD(z_{test}) - MD_{0}(z_{test})
\end{equation} \normalsize
where $MD_0$ represents the distance from a sample to the global distribution \cite{ren2022out}. MD is becoming popular as a new concept and has extended with multiple variants \cite{vazhentsev-etal-2023-hybrid,ren2023outofdistributiondetectionselectivegeneration,lee2018simpleunifiedframeworkdetecting}.

% Moreover, OOD detection techniques can be extended to address dataset shift \cite{NEURIPS2018_abdeb6f5}, as both tasks focus on scenarios that deviate from the training data. \citet{NEURIPS2019_8558cb40} explored predictive uncertainty in the presence of dataset shift, providing an important baseline for subsequent research in this domain.

\subsection{Data Annotation}
\label{sec:dataAnnotation}
\paragraph{Mitigate the Randomness Introduced by Annotators}
Supervised learning fundamentally depends on manually labeled data, often referred to as “gold standard” annotations. However, human annotators are inherently susceptible to variability and subjective interpretation \cite{zhu2023investigating}, contributing to the aleatoric uncertainty.

\citet{zhu2023investigating} proposed a method using BNNs to detect annotation errors. In recognition of human cognitive bias, the authors introduced a non-zero value to represent the variance associated with these errors. Similarly, \citet{Ge_Hu_Ma_Liu_Zhang_2024} applied Monte Carlo Dropout (MCD) to balance annotation errors during the distillation process across multiple languages. Beyond error detection, uncertainty plays a critical role in other aspects of  Named Entity Recognition (NER). Considering that NER is extremely sensitive to “gold standard” data \citet{he2024uncertaintyestimationsequentiallabeling}, recognizing and utilizing the data uncertainty plays a critical role in boosting its performance \citet{he2024uncertaintyestimationsequentiallabeling,liu2022uamner}. 
% For instance, \citet{he2024uncertaintyestimationsequentiallabeling} incorporated self-attention blocks into the network architecture, allowing the model to estimate annotation uncertainty while accounting for the uncertainty of neighboring tokens. \citet{liu2022uamner} presented UAMNer, a multi-model NER method that initially applies text-only NER to generate multiple candidate outputs. The model then refines these results using a multi-modal transformer that incorporates image data. This approach effectively leverages uncertainty in the multi-modal domain.

Furthermore, uncertainty can be integrated with LLMs.\citet{10.1145/3589334.3645414} introduced a method linking NER with ICL. In their approach, several smaller models first perform traditional NER. Then, the result with the lowest uncertainty is selected and incorporated into the prompt for ICL. 

\paragraph{Active Data Annotation}
Active Learning (AL) \cite{cohn1996active} aims to enhance data labeling efficiency by identifying the most informative unlabeled data for annotation within reasonable budget constraints. Uncertainty can serve as a key metric for determining which data points are most valuable to annotate.

For traditional supervised active learning, uncertainty sampling is widely considered one of the most effective approaches \cite{shen2017deep,schroder2021small,margatina2021importance}. However, in ICL scenarios, researchers have reported mixed results. For instance, \citet{diao2023active} demonstrated that selecting the most uncertain questions for annotation yielded promising results on eight widely used reasoning task datasets. In contrast, \citet{margatina2023active}  found that uncertainty-based methods underperformed compared to other prevalent AL algorithms.

A potential explanation for this discrepancy is that larger models benefit from demonstrations with higher uncertainty, whereas smaller models, such as GPT-2 and GPT-large, perform better when provided with low-uncertainty prompts.

\subsection{Question Clarification}
\label{sec:questionClarification}
In real-world scenarios, queries often contain some degree of ambiguity due to missing background knowledge, insufficient context, or open-endedness. 
%Uncertainty can help identify such ambiguous queries and guide appropriate follow-up actions.
Recently, several works have employed various uncertainty-based methods to identify ambiguous questions and guide appropriate follow-up actions.

% \citet{manggala2023aligning} investigate the alignment between predictive uncertainty and ambiguous instructions in visually grounded communication. They generate pairs of clear and ambiguous instructions through minimal edits, such as removing color or quantity information. If the predictive uncertainty for an instruction is significantly higher than for its clear counterpart, the instruction is deemed ambiguous, prompting the model to ask clarifying questions.

% \citet{hou2023decomposing} propose an input clarification ensembling framework that generates multiple input clarifications. By evaluating the disagreement among these clarifications, they quantify aleatoric uncertainty arising from input ambiguity. Similarly, \citet{cole2023selectively} measure denotational uncertainty—defined as question ambiguity—by generating various interpretations of a question and selecting the most likely one.

% Additionally, \citet{yona2024narrowing} introduce the GRANOLA QA evaluation method, which allows for multiple levels of granularity in correct answers. Their decoding strategy, Decoding with Response Aggregation (DRAG), adjusts answer granularity based on the model's uncertainty. When uncertain about a specific answer, the model may provide a coarser-grained response that, while less informative, has a higher likelihood of being correct.

\citet{manggala2023aligning} investigate the alignment between predictive uncertainty and ambiguous instructions in visually grounded communication. Specifically, they generated pairs of clear and ambiguous instructions through minimal edits (such as removing color or quantity information). Suppose the predictive uncertainty for an instruction is significantly higher than its clear counterpart. In that case, the instruction can be deemed ambiguous, at which point the model should ask clarifying questions to resolve the ambiguity.

\citet{hou2023decomposing} proposes an input clarification ensembling framework that generates an ensemble of input clarifications. By evaluating the disagreement among these clarifications, the authors quantify aleatoric uncertainty (due to input ambiguity) arising from total ambiguity. Similarly, \citet{cole2023selectively} measures denotational uncertainty—defined as question ambiguity—by generating various possible interpretations of a question and selecting the most likely interpretation.

On a different note, \citet{yona2024narrowing} proposes the GRANOLA QA evaluation method, which allows multiple levels of granularity for correct answers. They introduced a new decoding strategy, Decoding with Response Aggregation (DRAG), which adjusts the granularity of answers based on the model's level of uncertainty. If the model is uncertain about a specific answer, it may provide a coarser-grained response, which, while less informative, has a higher likelihood of being correct.

\section{Conclusion}
In this paper, we provide a comprehensive survey of uncertainty and its estimation from four distinct perspectives, offering valuable insights into uncertainty estimation in LLMs. Our work aims to bridge the gap between theoretical foundations and practical methodologies, contributing to a deeper understanding of uncertainty estimation in LLMs. Additionally, we showcase various methods for integrating uncertainty into a range of applications. We hope to inspire the development of innovative approaches that enhance the reliability of LLMs across diverse contexts.
% In this paper, we provide a comprehensive survey of uncertainty and explore its estimation from four distinct perspectives, offering valuable insights into uncertainty estimation in LLMs.
% : Bayesian inference (Section \ref{sec: BP}), information theory (Section \ref{sec: Info}), ensemble strategies (Section \ref{sec: Ensemble}), and language expression (Section \ref{sec: Language}).
% Our work aims to bridge the gap between theoretical foundations and practical methods, hoping to inspire the development of novel methods that can improve the reliability of LLMs in various applications. 

% Future research may focus on integrating these perspectives to create more robust frameworks for uncertainty estimation, ultimately contributing to more trustworthy AI systems.

% Future work should focus on integrating multiple theoretical perspectives—such as Bayesian inference, ensemble learning, and information theory—to create more adaptive and reliable frameworks. Additionally, applying these techniques in real-world scenarios, including out-of-distribution detection and data annotation, will enhance the utility of LLMs across various domains.

\section*{Limitations}
This review, while comprehensive, has certain limitations. First, the focus on four theoretical perspectives—Bayesian inference, information theory, ensemble strategies, and language expression—may overlook other emerging approaches that could contribute to uncertainty estimation. Furthermore, the rapidly evolving nature of the field means that new methodologies may emerge that are not covered in this review. Lastly, our survey and classification of methods is inherently subjective, as it is influenced by the selected literature and its interpretation.

\section*{Ethics Statement}

\paragraph{Use of AI Assistants}
We have employed ChatGPT as a writing assistant, primarily for polishing the text after the initial composition.

\bibliography{custom}

\begin{thebibliography}{80}
\providecommand{\natexlab}[1]{#1}

\bibitem[{Alkaissi and McFarlane(2023)}]{alkaissi2023artificial}
Hussam Alkaissi and Samy~I McFarlane. 2023.
\newblock Artificial hallucinations in chatgpt: implications in scientific writing.
\newblock \emph{Cureus}, 15(2).

\bibitem[{Arteaga et~al.(2024)Arteaga, Schön, and Pielawski}]{arteaga2024hallucinationdetectionllmsfast}
Gabriel~Y. Arteaga, Thomas~B. Schön, and Nicolas Pielawski. 2024.
\newblock \href {https://arxiv.org/abs/2409.02976} {Hallucination detection in llms: Fast and memory-efficient finetuned models}.
\newblock \emph{Preprint}, arXiv:2409.02976.

\bibitem[{Ash(1965)}]{Ashinformation}
Robert~B Ash. 1965.
\newblock \emph{Information theory}.
\newblock Dover Publications.

\bibitem[{Banerjee et~al.(2024)Banerjee, Agarwal, and Singla}]{banerjee2024llmshallucinateneedlive}
Sourav Banerjee, Ayushi Agarwal, and Saloni Singla. 2024.
\newblock \href {https://arxiv.org/abs/2409.05746} {Llms will always hallucinate, and we need to live with this}.
\newblock \emph{Preprint}, arXiv:2409.05746.

\bibitem[{Blundell et~al.(2015)Blundell, Cornebise, Kavukcuoglu, and Wierstra}]{blundell2015weight}
Charles Blundell, Julien Cornebise, Koray Kavukcuoglu, and Daan Wierstra. 2015.
\newblock Weight uncertainty in neural network.
\newblock In \emph{International conference on machine learning}, pages 1613--1622. PMLR.

\bibitem[{Chen and Mueller(2023)}]{chen2023quantifyinguncertaintyanswerslanguage}
Jiuhai Chen and Jonas Mueller. 2023.
\newblock \href {https://arxiv.org/abs/2308.16175} {Quantifying uncertainty in answers from any language model and enhancing their trustworthiness}.
\newblock \emph{Preprint}, arXiv:2308.16175.

\bibitem[{Chen and Mueller(2024)}]{chen-mueller-2024-quantifying}
Jiuhai Chen and Jonas Mueller. 2024.
\newblock \href {https://aclanthology.org/2024.acl-long.283} {Quantifying uncertainty in answers from any language model and enhancing their trustworthiness}.
\newblock In \emph{Proceedings of the 62nd Annual Meeting of the Association for Computational Linguistics (Volume 1: Long Papers)}, pages 5186--5200, Bangkok, Thailand. Association for Computational Linguistics.

\bibitem[{Cohn et~al.(1996)Cohn, Ghahramani, and Jordan}]{cohn1996active}
David~A Cohn, Zoubin Ghahramani, and Michael~I Jordan. 1996.
\newblock Active learning with statistical models.
\newblock \emph{Journal of artificial intelligence research}, 4:129--145.

\bibitem[{Cole et~al.(2023{\natexlab{a}})Cole, Zhang, Gillick, Eisenschlos, Dhingra, and Eisenstein}]{selectivelyanswering2023}
Jeremy~R. Cole, Michael J.~Q. Zhang, Daniel Gillick, Julian~Martin Eisenschlos, Bhuwan Dhingra, and Jacob Eisenstein. 2023{\natexlab{a}}.
\newblock \href {https://arxiv.org/abs/2305.14613} {Selectively answering ambiguous questions}.
\newblock \emph{Preprint}, arXiv:2305.14613.

\bibitem[{Cole et~al.(2023{\natexlab{b}})Cole, Zhang, Gillick, Eisenschlos, Dhingra, and Eisenstein}]{cole2023selectively}
Jeremy~R Cole, Michael~JQ Zhang, Daniel Gillick, Julian~Martin Eisenschlos, Bhuwan Dhingra, and Jacob Eisenstein. 2023{\natexlab{b}}.
\newblock Selectively answering ambiguous questions.
\newblock \emph{arXiv preprint arXiv:2305.14613}.

\bibitem[{Cosmides and Tooby(1996)}]{cosmides1996humans}
Leda Cosmides and John Tooby. 1996.
\newblock Are humans good intuitive statisticians after all? rethinking some conclusions from the literature on judgment under uncertainty.
\newblock \emph{cognition}, 58(1):1--73.

\bibitem[{Depeweg(2019)}]{Depeweg2019ModelingEA}
Stefan Depeweg. 2019.
\newblock \href {https://api.semanticscholar.org/CorpusID:208224498} {Modeling epistemic and aleatoric uncertainty with bayesian neural networks and latent variables}.

\bibitem[{Diao et~al.(2023)Diao, Wang, Lin, and Zhang}]{diao2023active}
Shizhe Diao, Pengcheng Wang, Yong Lin, and Tong Zhang. 2023.
\newblock Active prompting with chain-of-thought for large language models.
\newblock \emph{arXiv preprint arXiv:2302.12246}.

\bibitem[{Duan et~al.(2024)Duan, Zhang, Diffenderfer, Kailkhura, Sun, Stengel-Eskin, Bansal, Chen, and Xu}]{duan2024gtbench}
Jinhao Duan, Renming Zhang, James Diffenderfer, Bhavya Kailkhura, Lichao Sun, Elias Stengel-Eskin, Mohit Bansal, Tianlong Chen, and Kaidi Xu. 2024.
\newblock Gtbench: Uncovering the strategic reasoning limitations of llms via game-theoretic evaluations.
\newblock \emph{arXiv preprint arXiv:2402.12348}.

\bibitem[{Fadeeva et~al.(2023)Fadeeva, Vashurin, Tsvigun, Vazhentsev, Petrakov, Fedyanin, Vasilev, Goncharova, Panchenko, Panov et~al.}]{fadeeva2023lm}
Ekaterina Fadeeva, Roman Vashurin, Akim Tsvigun, Artem Vazhentsev, Sergey Petrakov, Kirill Fedyanin, Daniil Vasilev, Elizaveta Goncharova, Alexander Panchenko, Maxim Panov, et~al. 2023.
\newblock Lm-polygraph: Uncertainty estimation for language models.
\newblock \emph{arXiv preprint arXiv:2311.07383}.

\bibitem[{Fox(1980)}]{fox1980evolution}
Renee~C Fox. 1980.
\newblock The evolution of medical uncertainty.
\newblock \emph{The Milbank Memorial Fund Quarterly. Health and Society}, pages 1--49.

\bibitem[{Gal and Ghahramani(2016{\natexlab{a}})}]{pmlr-v48-gal16}
Yarin Gal and Zoubin Ghahramani. 2016{\natexlab{a}}.
\newblock \href {https://proceedings.mlr.press/v48/gal16.html} {Dropout as a bayesian approximation: Representing model uncertainty in deep learning}.
\newblock In \emph{Proceedings of The 33rd International Conference on Machine Learning}, volume~48 of \emph{Proceedings of Machine Learning Research}, pages 1050--1059, New York, New York, USA. PMLR.

\bibitem[{Gal and Ghahramani(2016{\natexlab{b}})}]{gal2016dropout}
Yarin Gal and Zoubin Ghahramani. 2016{\natexlab{b}}.
\newblock Dropout as a bayesian approximation: Representing model uncertainty in deep learning.
\newblock In \emph{international conference on machine learning}, pages 1050--1059. PMLR.

\bibitem[{Gawlikowski et~al.(2023)Gawlikowski, Tassi, Ali, Lee, Humt, Feng, Kruspe, Triebel, Jung, Roscher et~al.}]{gawlikowski2023survey}
Jakob Gawlikowski, Cedrique Rovile~Njieutcheu Tassi, Mohsin Ali, Jongseok Lee, Matthias Humt, Jianxiang Feng, Anna Kruspe, Rudolph Triebel, Peter Jung, Ribana Roscher, et~al. 2023.
\newblock A survey of uncertainty in deep neural networks.
\newblock \emph{Artificial Intelligence Review}, 56(Suppl 1):1513--1589.

\bibitem[{Ge et~al.(2024)Ge, Hu, Ma, Liu, and Zhang}]{Ge_Hu_Ma_Liu_Zhang_2024}
Ling Ge, Chunming Hu, Guanghui Ma, Jihong Liu, and Hong Zhang. 2024.
\newblock \href {https://doi.org/10.1609/aaai.v38i16.29762} {Discrepancy and uncertainty aware denoising knowledge distillation for zero-shot cross-lingual named entity recognition}.
\newblock \emph{Proceedings of the AAAI Conference on Artificial Intelligence}, 38(16):18056--18064.

\bibitem[{Geng et~al.(2024)Geng, Cai, Wang, Koeppl, Nakov, and Gurevych}]{geng-etal-2024-survey}
Jiahui Geng, Fengyu Cai, Yuxia Wang, Heinz Koeppl, Preslav Nakov, and Iryna Gurevych. 2024.
\newblock \href {https://doi.org/10.18653/v1/2024.naacl-long.366} {A survey of confidence estimation and calibration in large language models}.
\newblock In \emph{Proceedings of the 2024 Conference of the North American Chapter of the Association for Computational Linguistics: Human Language Technologies (Volume 1: Long Papers)}, pages 6577--6595, Mexico City, Mexico. Association for Computational Linguistics.

\bibitem[{Graves(2011)}]{graves2011practical}
Alex Graves. 2011.
\newblock Practical variational inference for neural networks.
\newblock \emph{Advances in neural information processing systems}, 24.

\bibitem[{Groot and Valdenegro-Toro(2024)}]{groot2024overconfidencekeyverbalizeduncertainty}
Tobias Groot and Matias Valdenegro-Toro. 2024.
\newblock \href {https://arxiv.org/abs/2405.02917} {Overconfidence is key: Verbalized uncertainty evaluation in large language and vision-language models}.
\newblock \emph{Preprint}, arXiv:2405.02917.

\bibitem[{Hafner et~al.(2019)Hafner, Tran, Lillicrap, Irpan, and Davidson}]{hafner2019noisecontrastivepriorsfunctional}
Danijar Hafner, Dustin Tran, Timothy Lillicrap, Alex Irpan, and James Davidson. 2019.
\newblock \href {https://arxiv.org/abs/1807.09289} {Noise contrastive priors for functional uncertainty}.
\newblock \emph{Preprint}, arXiv:1807.09289.

\bibitem[{He et~al.(2024)He, Yu, Lei, Lu, and Chen}]{he2024uncertaintyestimationsequentiallabeling}
Jianfeng He, Linlin Yu, Shuo Lei, Chang-Tien Lu, and Feng Chen. 2024.
\newblock \href {https://arxiv.org/abs/2311.08726} {Uncertainty estimation on sequential labeling via uncertainty transmission}.
\newblock \emph{Preprint}, arXiv:2311.08726.

\bibitem[{Hou et~al.(2023)Hou, Liu, Qian, Andreas, Chang, and Zhang}]{hou2023decomposing}
Bairu Hou, Yujian Liu, Kaizhi Qian, Jacob Andreas, Shiyu Chang, and Yang Zhang. 2023.
\newblock Decomposing uncertainty for large language models through input clarification ensembling.
\newblock \emph{arXiv preprint arXiv:2311.08718}.

\bibitem[{Huang et~al.(2024)Huang, Wu, Yang, Zhang, and Wu}]{huang2024uncttp}
Hsiu-Yuan Huang, Zichen Wu, Yutong Yang, Junzhao Zhang, and Yunfang Wu. 2024.
\newblock \href {https://arxiv.org/abs/2408.09172} {Unc-ttp: A method for classifying llm uncertainty to improve in-context example selection}.
\newblock \emph{Preprint}, arXiv:2408.09172.

\bibitem[{Huang et~al.(2023)Huang, Ruan, Huang, Jin, Dong, Wu, Bensalem, Mu, Qi, Zhao, Cai, Zhang, Wu, Xu, Wu, Freitas, and Mustafa}]{huang2023survey}
Xiaowei Huang, Wenjie Ruan, Wei Huang, Gaojie Jin, Yi~Dong, Changshun Wu, Saddek Bensalem, Ronghui Mu, Yi~Qi, Xingyu Zhao, Kaiwen Cai, Yanghao Zhang, Sihao Wu, Peipei Xu, Dengyu Wu, Andre Freitas, and Mustafa~A. Mustafa. 2023.
\newblock \href {https://arxiv.org/abs/2305.11391} {A survey of safety and trustworthiness of large language models through the lens of verification and validation}.
\newblock \emph{Preprint}, arXiv:2305.11391.

\bibitem[{Hüllermeier and Waegeman(2021)}]{hullermeier_aleatoric_2021}
Eyke Hüllermeier and Willem Waegeman. 2021.
\newblock \href {https://doi.org/10.1007/s10994-021-05946-3} {Aleatoric and epistemic uncertainty in machine learning: an introduction to concepts and methods}.
\newblock \emph{Machine Learning}, 110(3):457--506.

\bibitem[{Iwata and Ghahramani(2017)}]{iwata2017improvingoutputuncertaintyestimation}
Tomoharu Iwata and Zoubin Ghahramani. 2017.
\newblock \href {https://arxiv.org/abs/1707.05922} {Improving output uncertainty estimation and generalization in deep learning via neural network gaussian processes}.
\newblock \emph{Preprint}, arXiv:1707.05922.

\bibitem[{Jordan et~al.(1999)Jordan, Ghahramani, Jaakkola, and Saul}]{jordan1999introduction}
Michael~I Jordan, Zoubin Ghahramani, Tommi~S Jaakkola, and Lawrence~K Saul. 1999.
\newblock An introduction to variational methods for graphical models.
\newblock \emph{Machine learning}, 37:183--233.

\bibitem[{Kadavath et~al.(2022)Kadavath, Conerly, Askell, Henighan, Drain, Perez, Schiefer, Hatfield-Dodds, DasSarma, Tran-Johnson et~al.}]{kadavath2022language}
Saurav Kadavath, Tom Conerly, Amanda Askell, Tom Henighan, Dawn Drain, Ethan Perez, Nicholas Schiefer, Zac Hatfield-Dodds, Nova DasSarma, Eli Tran-Johnson, et~al. 2022.
\newblock Language models (mostly) know what they know.
\newblock \emph{arXiv preprint arXiv:2207.05221}.

\bibitem[{Kiureghian and Ditlevsen(2009)}]{AleatoryOrEpistemic}
Armen~Der Kiureghian and Ove Ditlevsen. 2009.
\newblock \href {https://doi.org/10.1016/j.strusafe.2008.06.020} {Aleatory or epistemic? does it matter?}
\newblock \emph{Structural Safety}, 31(2):105--112.
\newblock Risk Acceptance and Risk Communication.

\bibitem[{Kojima et~al.(2022)Kojima, Gu, Reid, Matsuo, and Iwasawa}]{ZeroShot_CoT2022}
Takeshi Kojima, Shixiang~(Shane) Gu, Machel Reid, Yutaka Matsuo, and Yusuke Iwasawa. 2022.
\newblock \href {https://proceedings.neurips.cc/paper_files/paper/2022/file/8bb0d291acd4acf06ef112099c16f326-Paper-Conference.pdf} {Large language models are zero-shot reasoners}.
\newblock In \emph{Advances in Neural Information Processing Systems}, volume~35, pages 22199--22213. Curran Associates, Inc.

\bibitem[{Kuhn et~al.(2023)Kuhn, Gal, and Farquhar}]{kuhn2023semantic}
Lorenz Kuhn, Yarin Gal, and Sebastian Farquhar. 2023.
\newblock Semantic uncertainty: Linguistic invariances for uncertainty estimation in natural language generation.
\newblock \emph{arXiv preprint arXiv:2302.09664}.

\bibitem[{Kullback and Leibler(1951)}]{1320776d-9e76-337e-a755-73010b6e4b64}
S.~Kullback and R.~A. Leibler. 1951.
\newblock \href {http://www.jstor.org/stable/2236703} {On information and sufficiency}.
\newblock \emph{The Annals of Mathematical Statistics}, 22(1):79--86.

\bibitem[{Lakshminarayanan et~al.(2017)Lakshminarayanan, Pritzel, and Blundell}]{lakshminarayanan2017simple}
Balaji Lakshminarayanan, Alexander Pritzel, and Charles Blundell. 2017.
\newblock Simple and scalable predictive uncertainty estimation using deep ensembles.
\newblock \emph{Advances in neural information processing systems}, 30.

\bibitem[{Lambert et~al.(2024)Lambert, Forbes, Doyle, Dehaene, and Dojat}]{LAMBERT2024102830}
Benjamin Lambert, Florence Forbes, Senan Doyle, Harmonie Dehaene, and Michel Dojat. 2024.
\newblock \href {https://doi.org/10.1016/j.artmed.2024.102830} {Trustworthy clinical ai solutions: A unified review of uncertainty quantification in deep learning models for medical image analysis}.
\newblock \emph{Artificial Intelligence in Medicine}, 150:102830.

\bibitem[{Lee et~al.(2018{\natexlab{a}})Lee, Lee, Lee, and Shin}]{NEURIPS2018_abdeb6f5}
Kimin Lee, Kibok Lee, Honglak Lee, and Jinwoo Shin. 2018{\natexlab{a}}.
\newblock \href {https://proceedings.neurips.cc/paper_files/paper/2018/file/abdeb6f575ac5c6676b747bca8d09cc2-Paper.pdf} {A simple unified framework for detecting out-of-distribution samples and adversarial attacks}.
\newblock In \emph{Advances in Neural Information Processing Systems}, volume~31. Curran Associates, Inc.

\bibitem[{Lee et~al.(2018{\natexlab{b}})Lee, Lee, Lee, and Shin}]{lee2018simpleunifiedframeworkdetecting}
Kimin Lee, Kibok Lee, Honglak Lee, and Jinwoo Shin. 2018{\natexlab{b}}.
\newblock \href {https://arxiv.org/abs/1807.03888} {A simple unified framework for detecting out-of-distribution samples and adversarial attacks}.
\newblock \emph{Preprint}, arXiv:1807.03888.

\bibitem[{Li et~al.(2023)Li, Cheng, Zhao, Nie, and Wen}]{li-etal-2023-halueval}
Junyi Li, Xiaoxue Cheng, Xin Zhao, Jian-Yun Nie, and Ji-Rong Wen. 2023.
\newblock \href {https://doi.org/10.18653/v1/2023.emnlp-main.397} {{H}alu{E}val: A large-scale hallucination evaluation benchmark for large language models}.
\newblock In \emph{Proceedings of the 2023 Conference on Empirical Methods in Natural Language Processing}, pages 6449--6464, Singapore. Association for Computational Linguistics.

\bibitem[{Lin et~al.(2022)Lin, Hilton, and Evans}]{lin2022teachingmodelsexpressuncertainty}
Stephanie Lin, Jacob Hilton, and Owain Evans. 2022.
\newblock \href {https://arxiv.org/abs/2205.14334} {Teaching models to express their uncertainty in words}.
\newblock \emph{Preprint}, arXiv:2205.14334.

\bibitem[{Lin et~al.(2024)Lin, Trivedi, and Sun}]{lin2024generatingconfidenceuncertaintyquantification}
Zhen Lin, Shubhendu Trivedi, and Jimeng Sun. 2024.
\newblock \href {https://arxiv.org/abs/2305.19187} {Generating with confidence: Uncertainty quantification for black-box large language models}.
\newblock \emph{Preprint}, arXiv:2305.19187.

\bibitem[{Liu et~al.(2020)Liu, Lin, Padhy, Tran, Bedrax-Weiss, and Lakshminarayanan}]{liu2020simpleprincipleduncertaintyestimation}
Jeremiah~Zhe Liu, Zi~Lin, Shreyas Padhy, Dustin Tran, Tania Bedrax-Weiss, and Balaji Lakshminarayanan. 2020.
\newblock \href {https://arxiv.org/abs/2006.10108} {Simple and principled uncertainty estimation with deterministic deep learning via distance awareness}.
\newblock \emph{Preprint}, arXiv:2006.10108.

\bibitem[{Liu et~al.(2022)Liu, Wang, Zhang, Qing, and He}]{liu2022uamner}
Luping Liu, Meiling Wang, Mozhi Zhang, Linbo Qing, and Xiaohai He. 2022.
\newblock Uamner: uncertainty-aware multimodal named entity recognition in social media posts.
\newblock \emph{Applied Intelligence}, 52(4):4109--4125.

\bibitem[{Malinin(2019)}]{malinin2019uncertainty}
Andrey Malinin. 2019.
\newblock \emph{Uncertainty estimation in deep learning with application to spoken language assessment}.
\newblock Ph.D. thesis, University of Cambridge.

\bibitem[{Malinin and Gales(2018)}]{malinin2018predictiveuncertaintyestimationprior}
Andrey Malinin and Mark Gales. 2018.
\newblock \href {https://arxiv.org/abs/1802.10501} {Predictive uncertainty estimation via prior networks}.
\newblock \emph{Preprint}, arXiv:1802.10501.

\bibitem[{Manakul et~al.(2023)Manakul, Liusie, and Gales}]{manakul2023selfcheckgpt}
Potsawee Manakul, Adian Liusie, and Mark J.~F. Gales. 2023.
\newblock \href {https://arxiv.org/abs/2303.08896} {Selfcheckgpt: Zero-resource black-box hallucination detection for generative large language models}.
\newblock \emph{Preprint}, arXiv:2303.08896.

\bibitem[{Manggala and Monz(2023)}]{manggala2023aligning}
Putra Manggala and Christof Monz. 2023.
\newblock Aligning predictive uncertainty with clarification questions in grounded dialog.
\newblock In \emph{Findings of the Association for Computational Linguistics: EMNLP 2023}, pages 14988--14998.

\bibitem[{Margatina et~al.(2021)Margatina, Barrault, and Aletras}]{margatina2021importance}
Katerina Margatina, Loic Barrault, and Nikolaos Aletras. 2021.
\newblock On the importance of effectively adapting pretrained language models for active learning.
\newblock \emph{arXiv preprint arXiv:2104.08320}.

\bibitem[{Margatina et~al.(2023)Margatina, Schick, Aletras, and Dwivedi-Yu}]{margatina2023active}
Katerina Margatina, Timo Schick, Nikolaos Aletras, and Jane Dwivedi-Yu. 2023.
\newblock \href {https://arxiv.org/abs/2305.14264} {Active learning principles for in-context learning with large language models}.
\newblock \emph{Preprint}, arXiv:2305.14264.

\bibitem[{Mora-Cross and Calderon-Ramirez(2024)}]{mora2024uncertainty}
Maria Mora-Cross and Saul Calderon-Ramirez. 2024.
\newblock Uncertainty estimation in large language models to support biodiversity conservation.
\newblock In \emph{Proceedings of the 2024 Conference of the North American Chapter of the Association for Computational Linguistics: Human Language Technologies (Volume 6: Industry Track)}, pages 368--378.

\bibitem[{Mukhoti et~al.(2023)Mukhoti, Kirsch, van Amersfoort, Torr, and Gal}]{Mukhoti_2023_CVPR}
Jishnu Mukhoti, Andreas Kirsch, Joost van Amersfoort, Philip~H.S. Torr, and Yarin Gal. 2023.
\newblock Deep deterministic uncertainty: A new simple baseline.
\newblock In \emph{Proceedings of the IEEE/CVF Conference on Computer Vision and Pattern Recognition (CVPR)}, pages 24384--24394.

\bibitem[{Ren et~al.(2022)Ren, Luo, Zhao, Krishna, Saleh, Lakshminarayanan, and Liu}]{ren2022out}
Jie Ren, Jiaming Luo, Yao Zhao, Kundan Krishna, Mohammad Saleh, Balaji Lakshminarayanan, and Peter~J Liu. 2022.
\newblock Out-of-distribution detection and selective generation for conditional language models.
\newblock In \emph{The Eleventh International Conference on Learning Representations}.

\bibitem[{Ren et~al.(2023)Ren, Luo, Zhao, Krishna, Saleh, Lakshminarayanan, and Liu}]{ren2023outofdistributiondetectionselectivegeneration}
Jie Ren, Jiaming Luo, Yao Zhao, Kundan Krishna, Mohammad Saleh, Balaji Lakshminarayanan, and Peter~J. Liu. 2023.
\newblock \href {https://arxiv.org/abs/2209.15558} {Out-of-distribution detection and selective generation for conditional language models}.
\newblock \emph{Preprint}, arXiv:2209.15558.

\bibitem[{Renda et~al.(2019)Renda, Barsacchi, Bechini, and Marcelloni}]{RENDA20191}
Alessandro Renda, Marco Barsacchi, Alessio Bechini, and Francesco Marcelloni. 2019.
\newblock \href {https://doi.org/10.1016/j.eswa.2019.06.025} {Comparing ensemble strategies for deep learning: An application to facial expression recognition}.
\newblock \emph{Expert Systems with Applications}, 136:1--11.

\bibitem[{Schr{\"o}der et~al.(2021)Schr{\"o}der, M{\"u}ller, Niekler, and Potthast}]{schroder2021small}
Christopher Schr{\"o}der, Lydia M{\"u}ller, Andreas Niekler, and Martin Potthast. 2021.
\newblock Small-text: Active learning for text classification in python.
\newblock \emph{arXiv preprint arXiv:2107.10314}.

\bibitem[{Shen et~al.(2023)Shen, Chen, Backes, and Zhang}]{shen2023chatgpt}
Xinyue Shen, Zeyuan Chen, Michael Backes, and Yang Zhang. 2023.
\newblock \href {https://arxiv.org/abs/2304.08979} {In chatgpt we trust? measuring and characterizing the reliability of chatgpt}.
\newblock \emph{Preprint}, arXiv:2304.08979.

\bibitem[{Shen et~al.(2017)Shen, Yun, Lipton, Kronrod, and Anandkumar}]{shen2017deep}
Yanyao Shen, Hyokun Yun, Zachary~C Lipton, Yakov Kronrod, and Animashree Anandkumar. 2017.
\newblock Deep active learning for named entity recognition.
\newblock \emph{arXiv preprint arXiv:1707.05928}.

\bibitem[{Shridhar et~al.(2019)Shridhar, Laumann, and Liwicki}]{shridhar2019comprehensiveguidebayesianconvolutional}
Kumar Shridhar, Felix Laumann, and Marcus Liwicki. 2019.
\newblock \href {https://arxiv.org/abs/1901.02731} {A comprehensive guide to bayesian convolutional neural network with variational inference}.
\newblock \emph{Preprint}, arXiv:1901.02731.

\bibitem[{Shrivastava et~al.(2023)Shrivastava, Liang, and Kumar}]{shrivastava2023llamas}
Vaishnavi Shrivastava, Percy Liang, and Ananya Kumar. 2023.
\newblock Llamas know what gpts don't show: Surrogate models for confidence estimation.
\newblock \emph{arXiv preprint arXiv:2311.08877}.

\bibitem[{Simpkin and Schwartzstein(2016)}]{simpkin2016tolerating}
A~Simpkin and Richard Schwartzstein. 2016.
\newblock Tolerating uncertainty—the next medical revolution?
\newblock \emph{New England Journal of Medicine}, 375(18).

\bibitem[{Srivastava et~al.(2014)Srivastava, Hinton, Krizhevsky, Sutskever, and Salakhutdinov}]{srivastava2014dropout}
Nitish Srivastava, Geoffrey Hinton, Alex Krizhevsky, Ilya Sutskever, and Ruslan Salakhutdinov. 2014.
\newblock Dropout: a simple way to prevent neural networks from overfitting.
\newblock \emph{The journal of machine learning research}, 15(1):1929--1958.

\bibitem[{Tian et~al.(2023)Tian, Mitchell, Zhou, Sharma, Rafailov, Yao, Finn, and Manning}]{tian2023justaskcalibrationstrategies}
Katherine Tian, Eric Mitchell, Allan Zhou, Archit Sharma, Rafael Rafailov, Huaxiu Yao, Chelsea Finn, and Christopher~D. Manning. 2023.
\newblock \href {https://arxiv.org/abs/2305.14975} {Just ask for calibration: Strategies for eliciting calibrated confidence scores from language models fine-tuned with human feedback}.
\newblock \emph{Preprint}, arXiv:2305.14975.

\bibitem[{Vazhentsev et~al.(2023)Vazhentsev, Kuzmin, Tsvigun, Panchenko, Panov, Burtsev, and Shelmanov}]{vazhentsev-etal-2023-hybrid}
Artem Vazhentsev, Gleb Kuzmin, Akim Tsvigun, Alexander Panchenko, Maxim Panov, Mikhail Burtsev, and Artem Shelmanov. 2023.
\newblock \href {https://doi.org/10.18653/v1/2023.acl-long.652} {Hybrid uncertainty quantification for selective text classification in ambiguous tasks}.
\newblock In \emph{Proceedings of the 61st Annual Meeting of the Association for Computational Linguistics (Volume 1: Long Papers)}, pages 11659--11681, Toronto, Canada. Association for Computational Linguistics.

\bibitem[{Wang et~al.(2022)Wang, Zhang, Deng, Gardner, Roth, and Chen}]{wang2022extractingguessingimprovingfaithfulness}
Haoyu Wang, Hongming Zhang, Yuqian Deng, Jacob~R. Gardner, Dan Roth, and Muhao Chen. 2022.
\newblock \href {https://arxiv.org/abs/2210.04992} {Extracting or guessing? improving faithfulness of event temporal relation extraction}.
\newblock \emph{Preprint}, arXiv:2210.04992.

\bibitem[{Wang et~al.(2023)Wang, Wei, Schuurmans, Le, Chi, Narang, Chowdhery, and Zhou}]{selfconsistency2023}
Xuezhi Wang, Jason Wei, Dale Schuurmans, Quoc Le, Ed~Chi, Sharan Narang, Aakanksha Chowdhery, and Denny Zhou. 2023.
\newblock \href {https://arxiv.org/abs/2203.11171} {Self-consistency improves chain of thought reasoning in language models}.
\newblock \emph{Preprint}, arXiv:2203.11171.

\bibitem[{Wen et~al.(2020)Wen, Tran, and Ba}]{wen2020batchensemblealternativeapproachefficient}
Yeming Wen, Dustin Tran, and Jimmy Ba. 2020.
\newblock \href {https://arxiv.org/abs/2002.06715} {Batchensemble: An alternative approach to efficient ensemble and lifelong learning}.
\newblock \emph{Preprint}, arXiv:2002.06715.

\bibitem[{Wimmer et~al.(2023)Wimmer, Sale, Hofman, Bischl, and H\"ullermeier}]{pmlr-v216-wimmer23a}
Lisa Wimmer, Yusuf Sale, Paul Hofman, Bernd Bischl, and Eyke H\"ullermeier. 2023.
\newblock \href {https://proceedings.mlr.press/v216/wimmer23a.html} {Quantifying aleatoric and epistemic uncertainty in machine learning: Are conditional entropy and mutual information appropriate measures?}
\newblock In \emph{Proceedings of the Thirty-Ninth Conference on Uncertainty in Artificial Intelligence}, volume 216 of \emph{Proceedings of Machine Learning Research}, pages 2282--2292. PMLR.

\bibitem[{Xiao and Wang(2018)}]{xiao2018quantifyinguncertaintiesnaturallanguage}
Yijun Xiao and William~Yang Wang. 2018.
\newblock \href {https://arxiv.org/abs/1811.07253} {Quantifying uncertainties in natural language processing tasks}.
\newblock \emph{Preprint}, arXiv:1811.07253.

\bibitem[{Xiao and Wang(2021)}]{xiao2021hallucinationpredictiveuncertaintyconditional}
Yijun Xiao and William~Yang Wang. 2021.
\newblock \href {https://arxiv.org/abs/2103.15025} {On hallucination and predictive uncertainty in conditional language generation}.
\newblock \emph{Preprint}, arXiv:2103.15025.

\bibitem[{Xiao et~al.(2022)Xiao, Liang, Bhatt, Neiswanger, Salakhutdinov, and Morency}]{xiao-etal-2022-uncertainty}
Yuxin Xiao, Paul~Pu Liang, Umang Bhatt, Willie Neiswanger, Ruslan Salakhutdinov, and Louis-Philippe Morency. 2022.
\newblock \href {https://doi.org/10.18653/v1/2022.findings-emnlp.538} {Uncertainty quantification with pre-trained language models: A large-scale empirical analysis}.
\newblock In \emph{Findings of the Association for Computational Linguistics: EMNLP 2022}, pages 7273--7284, Abu Dhabi, United Arab Emirates. Association for Computational Linguistics.

\bibitem[{Xiong et~al.(2024)Xiong, Hu, Lu, Li, Fu, He, and Hooi}]{xiong2024llmsexpressuncertaintyempirical}
Miao Xiong, Zhiyuan Hu, Xinyang Lu, Yifei Li, Jie Fu, Junxian He, and Bryan Hooi. 2024.
\newblock \href {https://arxiv.org/abs/2306.13063} {Can llms express their uncertainty? an empirical evaluation of confidence elicitation in llms}.
\newblock \emph{Preprint}, arXiv:2306.13063.

\bibitem[{Xu et~al.(2024)Xu, Zhu, Zhang, Ma, Fan, Chen, and Yu}]{xu2024rejection}
Hongshen Xu, Zichen Zhu, Situo Zhang, Da~Ma, Shuai Fan, Lu~Chen, and Kai Yu. 2024.
\newblock \href {https://arxiv.org/abs/2403.18349} {Rejection improves reliability: Training llms to refuse unknown questions using rl from knowledge feedback}.
\newblock \emph{Preprint}, arXiv:2403.18349.

\bibitem[{Yona et~al.(2024)Yona, Aharoni, and Geva}]{yona2024narrowing}
Gal Yona, Roee Aharoni, and Mor Geva. 2024.
\newblock Narrowing the knowledge evaluation gap: Open-domain question answering with multi-granularity answers.
\newblock \emph{arXiv preprint arXiv:2401.04695}.

\bibitem[{Zhang et~al.(2023{\natexlab{a}})Zhang, Li, Das, Malin, and Kumar}]{zhang-etal-2023-sac3}
Jiaxin Zhang, Zhuohang Li, Kamalika Das, Bradley Malin, and Sricharan Kumar. 2023{\natexlab{a}}.
\newblock \href {https://doi.org/10.18653/v1/2023.findings-emnlp.1032} {{SAC}$^3$: Reliable hallucination detection in black-box language models via semantic-aware cross-check consistency}.
\newblock In \emph{Findings of the Association for Computational Linguistics: EMNLP 2023}, pages 15445--15458, Singapore. Association for Computational Linguistics.

\bibitem[{Zhang et~al.(2024{\natexlab{a}})Zhang, Li, Das, Malin, and Kumar}]{zhang2024sac3reliablehallucinationdetection}
Jiaxin Zhang, Zhuohang Li, Kamalika Das, Bradley~A. Malin, and Sricharan Kumar. 2024{\natexlab{a}}.
\newblock \href {https://arxiv.org/abs/2311.01740} {Sac3: Reliable hallucination detection in black-box language models via semantic-aware cross-check consistency}.
\newblock \emph{Preprint}, arXiv:2311.01740.

\bibitem[{Zhang et~al.(2023{\natexlab{b}})Zhang, Li, Cui, Cai, Liu, Fu, Huang, Zhao, Zhang, Chen, Wang, Luu, Bi, Shi, and Shi}]{zhang2023sirenssongaiocean}
Yue Zhang, Yafu Li, Leyang Cui, Deng Cai, Lemao Liu, Tingchen Fu, Xinting Huang, Enbo Zhao, Yu~Zhang, Yulong Chen, Longyue Wang, Anh~Tuan Luu, Wei Bi, Freda Shi, and Shuming Shi. 2023{\natexlab{b}}.
\newblock \href {https://arxiv.org/abs/2309.01219} {Siren's song in the ai ocean: A survey on hallucination in large language models}.
\newblock \emph{Preprint}, arXiv:2309.01219.

\bibitem[{Zhang et~al.(2024{\natexlab{b}})Zhang, Zhao, Gao, and Hu}]{10.1145/3589334.3645414}
Zhen Zhang, Yuhua Zhao, Hang Gao, and Mengting Hu. 2024{\natexlab{b}}.
\newblock \href {https://doi.org/10.1145/3589334.3645414} {Linkner: Linking local named entity recognition models to large language models using uncertainty}.
\newblock In \emph{Proceedings of the ACM Web Conference 2024}, WWW '24, page 4047–4058, New York, NY, USA. Association for Computing Machinery.

\bibitem[{Zhu et~al.(2023)Zhu, Ye, Li, Zhang, and Wu}]{zhu2023investigating}
Yu~Zhu, Yingchun Ye, Mengyang Li, Ji~Zhang, and Ou~Wu. 2023.
\newblock Investigating annotation noise for named entity recognition.
\newblock \emph{Neural Computing and Applications}, 35(1):993--1007.

\end{thebibliography}

\appendix

% \section{Example Appendix}
% \label{sec:appendix}

% This is an appendix.

\end{document}